# Surface Defect Detection with Gabor Filter Using Reconstruction-Based Blurring U-Net-ViT

Jongwook Si[a], Sungyoung Kim[b]

[a]Dept. of Computer・AI Convergence Engineering, Kumoh National Institute of Technology, Gumi, Republic of Korea
[b]School of Computer Engineering, Kumoh National Institute of Technology, Gumi, Republic of Korea

**Abstract**

This paper proposes a novel approach to enhance the accuracy and reliability of texture-based surface defect detection using Gabor filters and a blurring U-Net-ViT model. By combining the local feature training of U-Net with the global processing of the Vision Transformer(ViT), the model effectively detects defects across various textures. A Gaussian filter-based loss function removes background noise and highlights defect patterns, while Salt-and-Pepper(SP) masking in the training process reinforces texture-defect boundaries, ensuring robust performance in noisy environments. Gabor filters are applied in post-processing to emphasize defect orientation and frequency characteristics. Parameter optimization, including filter size, sigma, wavelength, gamma, and orientation, maximizes performance across datasets like MVTec-AD, Surface Crack Detection, and Marble Surface Anomaly Dataset, achieving an average Area Under the Curve(AUC) of 0.939. The ablation studies validate that the optimal filter size and noise probability significantly enhance defect detection performance.

"Keywords: Texture; Defect Detection; Gabor Filter; Reconstruction; U-Net; Vision Transformer"

## 1. Introduction

In industrial processes, defect detection is an essential procedure for ensuring both the functional safety of products and structures and their structural integrity [1-4]. It is not only directly linked to a company's production efficiency but also serves as a key technology for preventing safety accidents and reducing post-quality-assurance costs. Although the concepts of Smart Factory and Digital Twin are expanding automation and intelligent monitoring in production lines [5-7], traditional manual inspection or simple image processing techniques inadequate in effectively classifying the diverse types of defects encountered in modern industrial environments [8-9].

In the past, surface conditions were primarily evaluated by combining expert manual inspection with basic image processing techniques. However, modern processes increasingly employ complex materials with highly diverse textures [10-12]. For example, polymer surfaces may have irregular distributions of fluorescent substances or adhesives, while metallic surfaces may exhibit varying patterns depending on the degree of reflection or gloss. These unpredictable factors make it difficult to achieve high-accuracy defect detection using conventional threshold-based analysis. With the rapid advancement of machine learning and deep learning technologies, automated defect detection research in nondestructive testing (NDT) contexts has been booming [13-16]. Particularly, Convolutional Neural Networks (CNNs) excel at extracting low-dimensional features—such as texture, luminance, and edges—from local regions and incrementally training higher-dimensional representations [17-18]. This has made CNNs highly applicable to various defect detection tasks. A representative example is U-Net, which leverages an Encoder-Decoder architecture with skip connections to comprehensively utilize multiscale features, making it well-suited for generating segmentation maps of defect regions. However, these CNN-based approaches typically require large, labeled datasets with explicit annotations of defect positions and types, prolonged training times driven by data characteristics, and show limitations in handling widely distributed defects or subtle noise in the image.

Recently, unsupervised learning-based reconstruction approaches have received significant attention in industrial defect detection [19-22]. In this approach, the model is trained using only normal (good) surface images, autonomously training the normal patterns. During inference, regions with large reconstruction errors are identified as defects. This approach offers substantial advantages in real-world contexts where defect types are varied, and labeled defect data are scarce, and can be broadly applied to surfaces such as tiles, roads, and leather. Moreover, because it primarily restores normal patterns, defects stand out more prominently even in noisy environments.

To address these needs, this paper proposes a hybrid system that combines U-Net with a Vision Transformer (ViT) [23] and applies Gabor filtering [24] as a post-processing step. U-Net excels at local receptive field filtering and multiscale feature extraction, effectively removing background noise with relatively fewer training epochs. Vision Transformer (ViT), on the other hand, conducts global information exchange via patch embedding and multi-head self-attention, making it advantageous in capturing widely distributed defect factors [25-26]. In this work, we integrate U-Net and ViT from an unsupervised learning perspective, focusing on reconstructing the "normal background" of the image. This process clarifies the distinction of defect patterns from the background, aided by the inclusion of blurring to reduce background noise and guide the model to concentrate solely on normal texture training. During inference, we further employ a Gabor filter to analyze high-frequency components that have specific wavelengths and orientations. Gabor filters emphasize edges or cracks [27-28] making defects more conspicuous in the reconstructed image. Through frequency-domain evaluation of discrepancies between the reconstructed and original images, regions with high defect probabilities are ultimately determined. This approach reduces the training time for U-Net-based defect detection models and provides flexibility in handling a variety of surfaces and noise conditions.

In conclusion, this study's defect detection system: Surface Defect Detection with Gabor Filter Using Reconstruction-Based Blurring U-Net–ViT contributes in three main ways:

**Proposing a Hybrid Network of U-Net and ViT for texture surface defect detection.** We combine the local filtering strengths of CNNs with the global interaction capabilities of Transformers to maximize noise removal and defect detection performance. This fusion enables fast and accurate restoration of complicated textures, making it suitable for achieving high-efficiency noise removal with minimal training time in real industrial processes.

**Background Noise Reduction and Model Stabilization via Blurring.** This study employs blurring during training to convert background information into low-frequency components, guiding the model to train the normal background patterns more smoothly and clearly. In addition, the Salt-and-Pepper (SP) masking technique is applied during training to promote stable learning and enhance model robustness.

**Defect Recognition Algorithm via Gabor Filter Post-Processing.** This paper applies Gabor filtering in the post-processing step to precisely extract high-frequency information remaining in reconstructed images. By re-evaluating defect likelihood in the frequency domain after background noise is removed, we achieve more accurate detection of various patterns, including fine cracks and point-shaped defects.

## 2. Related Works

This paper proposes a novel approach to texture-based surface defect detection by leveraging Gabor filters and a U-Net-ViT model. To establish the uniqueness of this study, we compare it with existing research, each possessing distinct technical strengths and limitations. Detailed analyses of the key related studies are presented below.

DAGAN [29] integrates dual autoencoders and Generative Adversarial Networks to enable anomaly detection and defect recognition. This method trains a reconstruction model based on normal data and identifies defect regions through reconstruction errors. While DAGAN [29] excels at detecting defects in relatively simple and uniform textures, its accuracy diminishes as texture complexity and irregularity increase. This limitation is particularly pronounced for periodic or high-frequency defects. Both DAGAN [29] and this study employ reconstruction-based methods for defect detection, focusing on highlighting defect regions indirectly. However, while DAGAN [29]'s dual autoencoder and GAN architecture is effective for simple textures, it struggles with more complex patterns. In contrast, this study combines ViT and U-Net to simultaneously train global and local features and incorporates SP masking to ensure robust defect detection in diverse environments. Additionally, the use of Gabor filters in the post-processing stage enhances the emphasis on defect frequency and directional characteristics.

V-DAFT [30] combines denoising autoencoders and Fourier transforms for defect detection in texture images. This method removes noise, restores global texture patterns, and highlights defect regions through Fourier analysis. V-DAFT [30] demonstrates strong performance on periodic textures, particularly when defects are concentrated in specific frequency bands. However, it is less effective for complex or irregular textures, as the noise removal process may blur the boundary between defects and the background. Both V-DAFT [30] and this study aim to enhance defect detection by removing noise and analyzing texture patterns. While V-DAFT [30] is effective at restoring global patterns using autoencoders and Fourier transforms, this study utilizes a Gaussian filter-based loss function to suppress background noise and incorporates SP Masking to maintain robust performance in complex environments. Unlike V-DAFT [30], which is primarily suited for periodic textures, this study achieves high accuracy even in irregular textures. F-GAN [31] leverages a GAN-based framework to fuse information across different frequency bands, optimizing defect detection performance. It effectively detects defects in high-frequency bands, emphasizing structural characteristics of defects. However, this approach is sensitive to the diversity of training data and may exhibit limited generalization across datasets. Additionally, its performance may decline when defects exist outside the high-frequency range. Both F-GAN [31] and this study focus on emphasizing defect frequency characteristics, with Gabor filters playing a critical role in both methods. Furthermore, both studies aim to improve detection performance by highlighting the local features of defects. However, this study integrates SP Masking into the training process, ensuring consistent performance across various environments.

DFR [32] proposes an unsupervised anomaly segmentation algorithm based on deep feature reconstruction. By training high-dimensional features of normal textures, DFR [32] identifies defect regions through reconstruction errors. This method is effective at capturing global texture patterns and achieves high accuracy when defects have distinct boundaries. However, its sensitivity to reconstruction errors diminishes when texture and defect boundaries are blurred or noisy. Both DFR [32] and this study utilize unsupervised learning to detect defects by training the features of normal data. While DFR [32] relies on reconstruction errors for defect identification, this study overcomes its limitations by combining Gaussian filters and SP Masking to emphasize texture and defect boundaries more effectively. The use of Gabor filters in post-processing further enhances defect detection reliability, distinguishing this study from DFR [32]. DifferNet [33] employs normalizing flows for semi-supervised defect detection, training the probability density of normal data and identifying anomalies as defects. DifferNet [33] performs well across various defect types and texture patterns, particularly in periodic patterns. However, its reliance on normal data limits its performance in cases where data diversity is low or feature differences between anomalies and normal data are minimal. Both DifferNet [33] and this study share the approach of identifying defects by training the probability density of normal data. While DifferNet [33] focuses on global texture patterns and performs strongly in periodic textures, it struggles when feature differences are subtle. This study, however, incorporates SP Masking training and Gabor filter-based post-processing to maintain high performance even under minimal texture differences. Furthermore, this study differentiates itself by integrating both global information and local defect features.

## 3. Proposed Frameworks for Defect Detection

The framework proposed in this paper is divided into two main processes: the training phase and the testing phase. During the training phase, the proposed Hybrid model is designed to operate in a reconstruction manner, ensuring that the input and output images maintain the same dimensions. While fundamentally based on the U-Net architecture, the model introduces a hybrid approach by integrating a ViT module before the bottleneck. This hybrid model is trained exclusively on normal data, adopting a patch-wise approach to split the input data into patches and train the inter-patch correlations. This enables the model to effectively account for both global and local information, allowing it to reconstruct an image that closely resembles the input normal data. A notable advantage of the proposed approach is that defect detection is not solely reliant on the model's training process. Instead, the trained model generates reconstructed outputs, which are subsequently analyzed using Gabor filtering for defect detection. The proposed overall framework is illustrated in Figure 1.

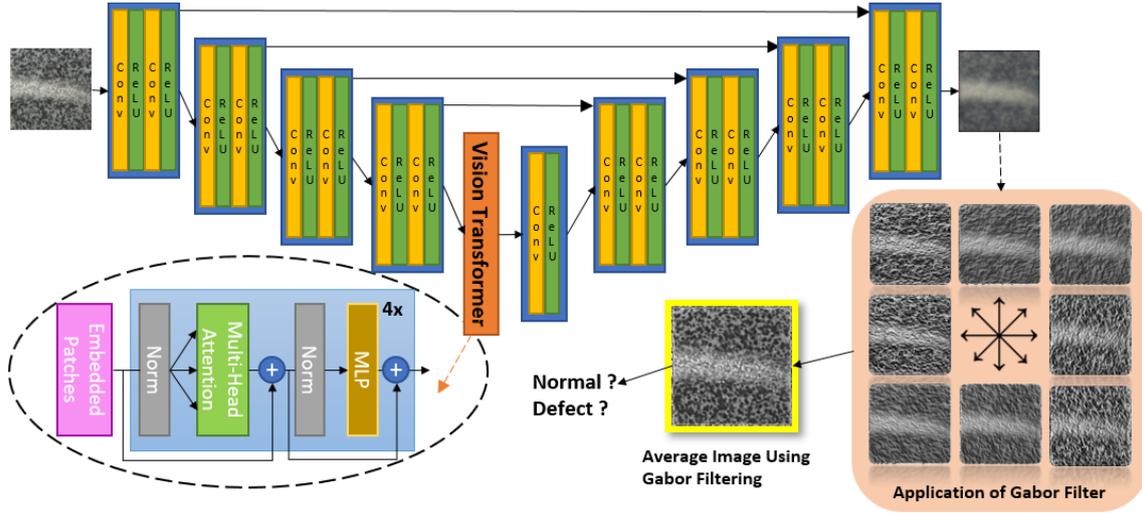

Fig. 1. Overall Framework for Surface Defect Detection Using Blurring U-Net-ViT and Gabor Filter

### 3.1 Design of Architecture

The proposed model is designed as a hybrid structure that combines the strengths of U-Net and ViT, designed for restoring noisy images and detecting defects. This model seamlessly integrates the local feature learning capability of the CNN-based U-Net and the global context learning capability of ViT, effectively handling various types of noise and defects in input data. To achieve this, the model is composed of four main components: Encoder, Vision Transformer Block, Bottleneck, and Decoder, with each stage working in a complementary manner.

The input to the model is an RGB image of size (N,3,256,256) noise. This input is passed to the first component of the model, the U-Net Encoder. The Encoder is designed based on CNNs and optimized for training local features. The first layer of the Encoder processes the input image using a 3×3 convolutional filter and a ReLU activation function. The convolution operation extracts low-level information, such as edges and textures, forming the initial stage of noise suppression. The ReLU activation function introduces nonlinearity, enhances the representational capacity of the model, and efficiently eliminates negative values. The output of the first layer is transformed into 64 feature maps, with dimensions (N,64,256,256). Subsequently, the MaxPooling operation reduces the resolution by half, changing the output tensor size to (N,64,128,128). MaxPooling helps retain essential image features while reducing computational costs and expanding the receptive field, enabling the training of broader information. The Encoder applies this structure repeatedly, increasing the number of filters at each layer. The number of filters is expanded as 64→128→256→512, and the spatial resolution gradually decreases during this process. The output of the final layer has a size of (N,512,32,32), containing the highest-level features. This abstract representation is a compressed version of the patterns trained by the Encoder and is passed to the Vision Transformer Block to train global interactions.

The hybrid structure combining U-Net and ViT is designed to simultaneously leverage both local features and global contextual information for texture-based defect detection. In this architecture, the encoder of the U-Net utilizes CNN-based blocks to hierarchically extract local features from the input image. Through this process, the final encoder layer generates a high-level semantic feature map with the shape (N,512,32,32), which is then passed to the ViT. First, during the Patch Embedding stage, the input tensor is divided into patches of size 16×16, and each patch is converted into a 512-dimensional embedding vector through Conv2D operations. Through this process, the tensor size changes from (N,512,32,32) to (N,4,512), where 4 represents the number of patches, and 512 is the embedding dimension for each patch. The Patch Embedding process compresses local information into a form suitable for global training using convolutional operations and ReLU activation functions. The core of ViT is the Multi-Head Self-Attention (MSA) mechanism. Self-Attention trains the relationships between patches by calculating attention weights using inputs transformed into Query (Q), Key (K), and Value (V). This process is expressed by the following equations, where α represents the attention weights that indicate the correlations between inputs. $d_k$ represents the dimensionality of the Key vector. This can be expressed mathematically as shown in (1). As a result, this hybrid architecture effectively combines the fine-grained local detail representation that CNNs excel at with the global contextual understanding offered by the Transformer. This enables both precise defect reconstruction and effective noise suppression. Particularly in cases involving subtle defects or complex texture structures, this approach demonstrates significantly more robust and reliable detection performance compared to conventional filter-based methods.

$$\text{Attention}(Q, K, V) = \text{softmax}\left((QK^{T})/\sqrt{(d_k)}\right) \cdot V \tag{1}$$

The final output of Self-Attention undergoes additional processing by a Feed-Forward Network (FFN). The FFN consists of two linear transformations and a GELU activation function, refining the representation of each patch. Residual connections ensure stable learning, and the output of the FFN retains its shape as (N,4,512). The final output of the ViT Block is reshaped to (N,512,2,2) to combine with the CNN Decoder.

The Bottleneck stage is responsible for training the most abstract features of the model, reducing the input tensor size from (N,512,2,2) to (N,1024,1,1). This stage consists of convolution and MaxPooling operations with ReLU activation functions. The Bottleneck generates high-level representations essential for noise removal and restoration, providing a foundation for the Decoder to begin the reconstruction process. The Decoder restores the resolution of the input image through upsampling and Skip Connections. In the upsampling stage, Interpolation operations double the tensor resolution, and Skip Connections utilize the low-level information trained by the Encoder. Each stage of the Decoder consists of convolution operations and ReLU activation functions. In the final stage, a 1×1 convolution generates the final RGB image. The final output has a size of (N,3,256,256), representing a clean image restored after noise removal.

## 3.2 Strategy of Training

The proposed model's training strategy includes partitioning data into a grid-based structure and randomly applying SP masking to specific regions to effectively perform noise removal and original image restoration. Unlike conventional methods that use only refined original images as input, this strategy is designed to enable the model to achieve robust performance under diverse data conditions. SP Masking and grid-based data augmentation allow the model to train both local and global features of the original data in a balanced manner, ensuring stable performance even in complex noise environments.

The training process begins by dividing the input image into a grid of size $k \times k$. The grid size $k$ can be adjusted based on the data resolution and the model's processing capability. This grid partitioning enables the model to train local and global relationships simultaneously. Each grid patch is processed independently, and SP Masking is randomly added to some of these patches. This diversifies the training data, allowing the model to adapt to various types of defects and noise conditions encountered in real-world scenarios. SP Masking operates by randomly modifying pixel values in each grid to either the maximum value 1 or the minimum value 0 based on a probabilistic distribution. $P_s$ is the probability of Salt masking, which sets the pixel value to 1, creating bright regions. Similarly, $P_p$ is the probability of Pepper masking, which sets the pixel value to 0, creating dark regions. SP Masking is applied to randomly selected patches within the grid-divided image, enabling the model to train under varied noise conditions within the same image. This process is applied uniformly to both the input image and the ground truth image. Here, $x$ represents the original image, and $SP(x)$ denotes the image with SP Masking applied. Consequently, the model is trained to restore the SP Masking-affected input image to closely resemble $x_i$, effectively training to recover the original data from noisy inputs.

### 3.3 Loss Functions

The proposed model's loss function is composed of a combination of Gaussian filtering-based loss and L1 loss, aimed at removing background noise and restoring the original image. Each loss function helps the model restore the low-frequency components of the input image and maintain overall pixel similarity. This design of the loss function is particularly effective for processing data containing SP masking, aiding in generating noise-free output images.

The loss function aims to restore images with SP Masking back to the original image. During the training process, the model trains to reconstruct the input image with added SP Masking so that it closely resembles the original image with SP Masking. To achieve this, a loss function combining two main components is proposed.

First, the L1 loss trains the model to maintain overall similarity between the reconstructed image and the original image. It calculates the absolute difference of each pixel, guiding the model to align the overall structure and details with the original data. The L1 loss is defined as follows in (2):

$$\mathcal{L}_{L1} = \frac{1}{N}\sum_{i=1}^{N}|\hat{x}_i - x_i| \tag{2}$$

Here, $\hat{x}_i$ represents the pixel value of the reconstructed image, and $x_i$ represents the pixel value of the original image. The L1 loss is a simple yet effective loss function that ensures the model retains the overall characteristics of the original data and plays a critical role in guaranteeing reconstruction quality.

Second, the Gaussian filtering-based loss emphasizes restoring low-frequency components, helping the model remove background noise and accurately train normal patterns. This loss is calculated $SP(x_i)$ and $SP(\hat{x}_i)$ by applying a Gaussian filter to both the input image and the reconstructed image, then measuring the difference between them. The Gaussian filter is designed with a kernel size of 11 and a standard deviation of 5 to stably extract low-frequency components over a wide area. The Gaussian filtering-based loss is defined as follows in (3), where $G$ denotes the result of applying a Gaussian filter to the input. And, $*$ denotes the 2D convolution operation. The Gaussian filtering-based loss encourages the model to accurately restore the low-frequency components of the background, contributing to effectively suppressing high-frequency components such as SP Masking. The final loss function is defined by combining the two losses, enabling the model to achieve both objectives in a balanced manner. It is expressed as follows in (4).

$$\mathcal{L}_{Gau} = \frac{1}{N}\sum_{i=1}^{N}|G*\hat{x}_i - G*x_i| \tag{3}$$

$$\mathcal{L}_{total} = \mathcal{L}_{L1} + \mathcal{L}_{Gau} \tag{4}$$

3.4   Defect Detection Using Gabor Filter

The defect detection process using Gabor filters emphasizes defect regions based on reconstructed images, setting the entire image as the detection area to effectively extract the location and characteristics of defects. This process involves applying Gabor filters to the reconstructed image, generating an average image, and calculating defect scores. Specifically, the reconstructed image undergoes a reconstruction process to remove background noise, thereby maximizing the sensitivity of the Gabor filter. It employed in this study is configured in 8 directions to emphasize defect textures, with each direction $\theta_k$ defined by (5):

$$\theta_k = \frac{\pi}{8}k + \frac{\pi}{16}, \quad k = 0, 1, \ldots, 7 \tag{5}$$

Here, the offset π/16 increases the precision of the filter direction, enabling more accurate capture of various defect orientations and texture variations. This is designed to comprehensively detect diverse defect textures. The mathematical expression of the Gabor filter is given in Equation (7), where $x'$ and $y'$ represent the coordinates rotated by the filter direction $\theta$, defined as shown in (6). Here, this means $x' = x\cos\theta + y\sin\theta$ and $y' = -x\sin\theta + y\cos\theta$.

$$G(x, y; \theta, \sigma, \lambda, \gamma) = \exp\left(-\frac{x'^2}{2\sigma^2} - \frac{\gamma^2 y'^2}{2\sigma^2}\right)\cos\left(2\pi\frac{x'}{\lambda}\right) \tag{6}$$

The Gabor filter includes key parameters: λ(wavelength), θ(orientation), σ(standard deviation), and γ(spatial aspect ratio). Each parameter plays a crucial role in defect detection. The parameter λ determines the frequency to which the filter responds, with larger values being sensitive to low-frequency components and smaller values to high-frequency components. This parameter is adjusted based on the size of the defect. The parameter σ controls the width of the Gaussian distribution of the filter, with larger values capturing wider defect areas and smaller values emphasizing localized defects. The parameter γ represents the spread ratio along the $x$- and $y$- axes. Additionally, we also consider the filter size as a separate parameter. The filter size determines the spatial region of the image that the Gabor filter processes, serving as a critical parameter in applications such as defect detection. Larger filter sizes are advantageous for analyzing broad patterns or textures, while smaller filter sizes are more effective for detecting localized features. For the reconstructed image $\hat{x}_i$, the Gabor filtering result $R_k(x, y)$ in each direction $k$ is calculated as shown in (7). Here, $G_k$ represents the Gabor filter in the $k$-th direction.

$$R_k(x, y) = (G_k * \hat{x}_i)(x, y) \tag{7}$$

The filtering results highlight defect textures and high-frequency components in each direction. Based on the results filtered in 8 directions, an average image $A(x_i)$ is generated. The average image normalizes the Gabor filtering results at each pixel, emphasizing defect textures while minimizing the influence of background noise.

A defect score is calculated based on this average image, using its maximum value for evaluation. The defect score is defined as shown in (8). Here, $1\{R_k(x, y) > 0\}$ is an indicator function that returns 1 if $\{R_k(x, y) > 0\}$, and 0 otherwise.

$$\text{DFscore} = \max\left(\frac{\sum_{k=0}^{7} R_k(x, y)}{\sum_{k=0}^{7} 1\{R_k(x, y) > 0\}}\right) \tag{8}$$

## 4. Experiments and Results Analysis

4.1 Datasets and Environment Settings

In this study, the performance of the Gabor filter-based defect detection algorithm was evaluated using various texture-based datasets. The experiments were conducted using three texture datasets, comprising a total of seven categories. The detailed numbers of data used for training and testing are presented in Table 1.

Table 1. Detailed Number of Used Data

| Dataset | Category | Train | Test | |
|---|---|---|---|---|
| | | Normal | Normal | Defect |
| MVTec-AD[34] | Carpet | 280 | 84 | 40 |
| | Grid | 264 | 63 | 57 |
| | Leather | 245 | 96 | 92 |
| | Tile | 230 | 99 | 84 |
| | Wood | 247 | 57 | 30 |
| Surface Crack Detection[35] | Pave&Concrete | 7,500 | 750 | 1,500 |
| Marble Surface Anomaly Dataset[36] | Marble | 860 | 340 | 348 |

The MVTec AD [34] dataset consists of five texture categories: Carpet, Grid, Leather, Tile, and Wood. Each category includes normal images and various defect images. The Surface Crack Detection [35] dataset includes images of cracks, damage, and foreign objects on road pavements and concrete surfaces. This dataset comprises a total of 40,000 images, equally divided into 20,000 normal images and 20,000 defect images. To ensure a balanced evaluation and reduce redundancy caused by similar data samples, this dataset was undersampled. The Marble Surface Anomaly [36] dataset includes defect images of cracks, scratches, and stains on marble surfaces. While defect images for training are available in this dataset, they were excluded as the proposed method focuses solely on training from normal data.

The implementation and performance evaluation of the Gabor filter-based defect detection algorithm were conducted under the following experimental setup. The operating system used was Ubuntu 18.04 LTS, chosen for its stability and efficiency in handling GPU acceleration and large-scale data processing. The experiments were carried out using Python 3.9 as the programming language to implement the proposed methodology. The deep learning framework employed was PyTorch 1.15, which facilitated efficient tensor computations and GPU parallel processing. The hardware utilized for these experiments was an NVIDIA GeForce RTX 3090 GPU.

4.2 Design of Gabor Filters

In this paper, key parameters of the Gabor filter were systematically explored to determine the optimal configuration for each dataset. The grid size was fixed at 8x8 ($k$=8), with the noise probabilities $P_S, P_P$ set to 0.05. Gabor filter applications were based on the reconstruction, with the number of epochs set to 10 and batchsize set to 8 with Adam Optimizer.

The kernel size of the filter was examined within the range of [5, 7, 9, ..., 39], while σ was explored over [1, 2, ..., 20]. The λ ranged from [1, 2, ..., 20], and γ was analyzed within [0.2, 0.4, ..., 4.0]. Each parameter was adjusted iteratively to account for the unique texture characteristics and defect types present in the datasets. The optimization process focused on fine-tuning the Gabor filter parameters for each dataset, resulting in the highest performance configurations. These designed filters are visualized in Figure 2, where the columns represent filter orientations k ranging from 0 to 7, and the rows depict the filters tailored to each dataset: Carpet, Grid, Leather, Tile, Wood from the MVTec AD dataset, as well as the Surface Crack Detection dataset and the Marble Surface Anomaly dataset. By aligning the filter configurations with the distinct characteristics of each dataset, this approach ensured precise defect detection and enhanced the robustness of the proposed methodology. The visualized filters effectively highlight the adaptability of the Gabor filter parameters across varying textures and defect categories.

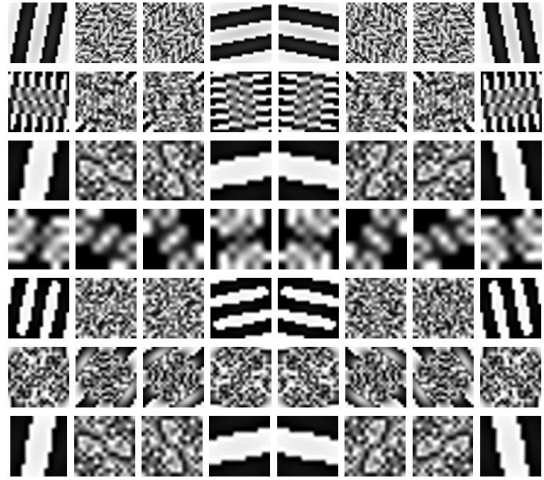

Fig. 2. Banks of Gabor Filters in Proposed Method.

In the MVTec-AD dataset, the Carpet category demonstrated optimal defect detection with a kernel size of 27, σ=12, λ=13, and γ=0.8. These parameter values effectively captured global texture variations while highlighting defect regions. The Grid category, characterized by periodic textures, required a narrower kernel size of 23, σ=7, and a shorter λ=4 to emphasize high-frequency discontinuities in the repetitive patterns. This configuration effectively detected defects such as line breaks. For the Leather category, which features fine, irregular textures, a kernel size of 15, σ=11, λ=13, and γ=1.8 were applied. These parameters accentuated specific directional features, enabling clear detection of scratches and dents. The Tile category, with its regular patterns and localized defects, was optimized using a smaller kernel size of 7, σ=2, and λ=5, allowing the filter to detect fine texture variations in the high-frequency domain. In the Wood category, which contains complex, repetitive patterns resembling wood grain, the parameters were set to a kernel size of 21, σ=14, λ=8, and γ=3. This configuration successfully balanced the detection of both global patterns and localized defects, such as disruptions in the grain structure.

For the Surface Crack Detection dataset, which focuses on irregular textures such as concrete surfaces, the kernel size was 21, σ=5, λ=10, and γ=1.2. This combination effectively reflected mid-frequency irregularities while capturing directional features to identify cracks and fractures. Similarly, the Marble Surface Anomaly dataset used the same configuration as the Leather category to emphasize fine defects such as cracks and stains.

4.3   Experiments

*4.3.1   Result Analysis*

The proposed Gabor filter-based defect detection algorithm was evaluated against the representative MVTec-AD dataset for surface texture anomaly detection. The comparison included state-of-the-art algorithms such as DAGAN[29], V-DAFT[30], F-GAN[31], DFR[32], and DifferNet[33], with performance assessed using the Area Under the Curve (AUC), a robust metric for quantifying the accuracy of defect detection. All comparative results reported in this study were sourced from the original publications. The performance evaluation results are shown in Table 2.

Table 2. Performance Comparisons with Related Works Using AUC(↑)

| Category | DAGAN [29] | V-DAFT [30] | F-GAN [31] | DFR [32] | DifferNet [33] | Ours |
|---|---|---|---|---|---|---|
| Carpet | 0.903 | 0.866 | 0.887 | **0.998** | 0.840 | 0.909 |
| Grid | 0.867 | 0.957 | 0.951 | 0.813 | 0.971 | **0.973** |
| Leather | 0.944 | 0.975 | 0.944 | 0.971 | **0.994** | 0.985 |
| Tile | 0.961 | 0.932 | 0.961 | 1.000 | 0.929 | 0.877 |
| Wood | 0.979 | 0.976 | 0.979 | 0.939 | **0.998** | 0.988 |
| Average AUC(↑) | 0.931 | 0.939 | 0.939 | 0.944 | **0.946** | **0.946** |

The Carpet category features relatively uniform textures with localized defects. Among the comparative methods, DAGAN achieved a high AUC of 0.903, indicating a strong performance in capturing these anomalies. V-DAFT and F-GAN recorded lower AUCs of 0.866 and 0.887, respectively, demonstrating limited capability in identifying localized defect textures. DFR achieved the highest AUC of 0.998, reflecting its effectiveness in reconstructing normal textures for enhanced defect detection. The Grid category is characterized by periodic, structured patterns, with defects manifesting as disruptions in these patterns. V-DAFT and F-GAN performed reliably with AUCs of 0.957 and 0.951, respectively, while DifferNet demonstrated the highest performance with an AUC of 0.971, underlining its proficiency in analyzing periodic patterns and detecting anomalies. In contrast, DAGAN achieved a relatively low AUC of 0.867, revealing its limitations in capturing periodic textural discontinuities. The proposed method outperformed all other algorithms with an AUC of 0.973, demonstrating that Gabor filters were particularly effective in enhancing the discontinuities in periodic Grid textures, making defect regions more prominent. Figure 3 provides a visual representation of the dataset and the outcomes of the proposed method. The first column illustrates the input images, while the second column shows the reconstruction results generated by the U-Net-ViT model. The third column displays the averaged Gabor-filtered images across eight directions, and columns 4 through 11 demonstrate the directional results for $k$=0 to $k$=7, providing detailed insights into the performance of Gabor filtering.

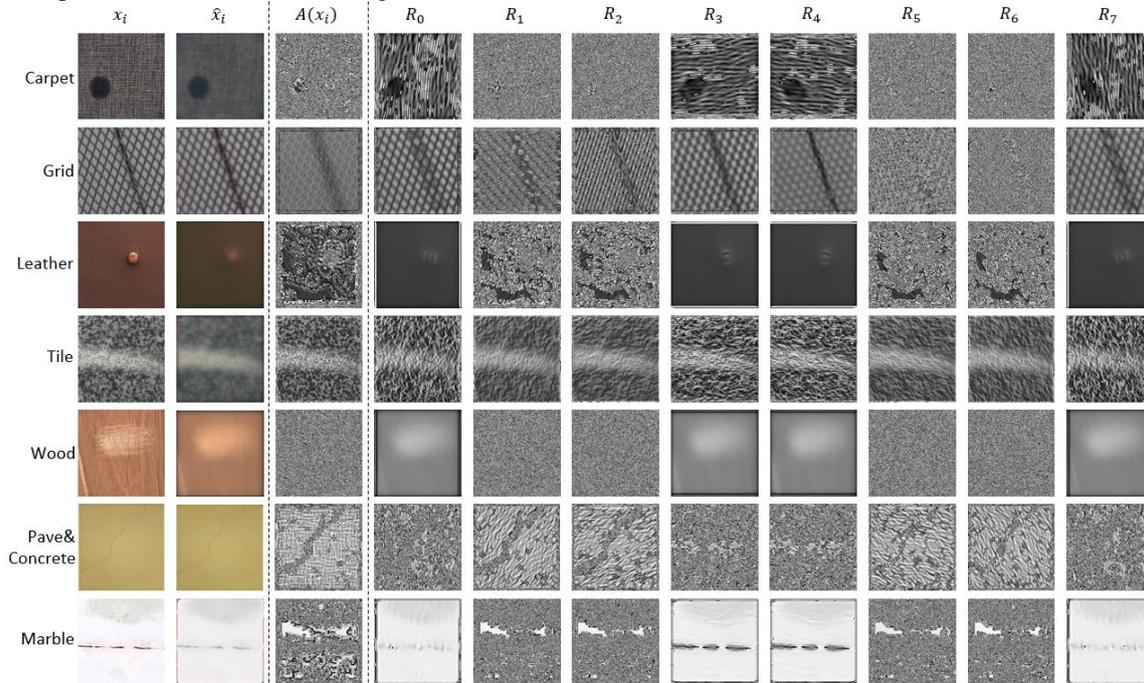

Fig. 3. The original image, the reconstructed result, the Gabor filter-applied result, and the direction-specific Gabor filters. ($x_i$: input image; $\hat{x}_i$: reconstructed image; $A(x_i)$: average $R_k$ image; $R_k$: image obtained by applying a Gabor filter in each direction)

These results illustrate the strengths of the proposed Gabor filter-based approach in leveraging texture-specific parameter optimization to enhance defect detection across various texture categories. By integrating the U-Net-ViT reconstruction model with Gabor filtering, the method not only improved defect detection accuracy but also demonstrated its robustness across different texture types, as validated by the performance metrics.

The Leather category is characterized by irregular textures and a relatively clean background with small defects. DAGAN and F-GAN both achieved an AUC of 0.944, indicating limited performance in handling the irregular texture of this category. V-DAFT demonstrated improved performance with an AUC of 0.975, while DifferNet excelled with an almost perfect AUC of 0.994. The proposed method achieved an AUC of 0.985, closely matching DifferNet's performance. This result highlights the effectiveness of Gabor filters in emphasizing defect textures in the Leather dataset, particularly in detecting small and irregular anomalies. The Tile category features symmetrical and regular textures with small localized defects. DAGAN and F-GAN both recorded an AUC of 0.961, demonstrating comparable performance. V-DAFT showed slightly lower performance with an AUC of 0.932. DFR achieved a perfect AUC of 1.000, reflecting its ability to accurately detect and restore defects in the regular texture of tiles. DifferNet achieved an AUC of 0.929, which was relatively lower. The proposed method, however, recorded an AUC of 0.877, indicating comparatively lower performance in this category. This result suggests that the sensitivity of Gabor filters to the symmetrical and regular textures of tiles might be slightly inferior to other algorithms. As observed in Figure 3, the tile patterns in the background became blurry during reconstruction but were partially restored by Gabor filtering, leading to some misclassifications. The Wood category contains textures of wood grain with localized defects, requiring the ability to handle various scales and texture patterns in defect detection. DAGAN and F-GAN both achieved an AUC of 0.979, with V-DAFT showing similar performance at an AUC of 0.976. DifferNet recorded the highest AUC of 0.998, demonstrating its ability to effectively train both the global wood grain texture and localized defects. The proposed method closely followed with an AUC of 0.988, indicating that Gabor filters were effective in emphasizing defect textures within the Wood dataset.

Analyzing the average AUC performance across all datasets, DAGAN, V-DAFT, and F-GAN showed similar levels of performance with AUCs of 0.931, 0.939, and 0.939, respectively. DFR improved upon this with an AUC of 0.944, while DifferNet achieved the highest average AUC of 0.946. The proposed method also achieved an average AUC of 0.946, matching DifferNet's performance. These results underscore the effectiveness of the Gabor filter-based post-processing approach in detecting defects across diverse texture categories, leveraging its ability to emphasize defect features while suppressing noise.

*4.3.2 Ablation Study*

To rigorously evaluate the contribution of the ViT module within the proposed Gabor-based hybrid architecture, we conducted an ablation study comparing performance with and without the ViT component. In this experiment, all other settings—including the Gabor filter preprocessing and U-Net structure—were kept identical, and only the ViT block was excluded to assess its isolated effect. The results are summarized in Table 3, where AUC (Area Under the Curve) values are reported for each texture category.

The average AUC across all seven texture categories, including five from MVTec AD (Carpet, Grid, Leather, Tile, Wood), and two from additional categories (Pave&Concrete and Marble), was 0.899 when using the U-Net only, and improved to 0.939 when the ViT module was integrated. This demonstrates a consistent and meaningful improvement in defect detection performance due to the inclusion of the ViT module.

A breakdown by category reveals the extent of this improvement. In the Carpet category, the AUC increased from 0.823 to 0.909, a numerical gain of 0.086, reflecting ViT's ability to enhance performance on textures that are highly irregular and lack strong repetitive structure. This improvement suggests that global contextual awareness from ViT is particularly effective in capturing spatial anomalies within unstructured textures. In the Grid category, the AUC rose from 0.886 to 0.973, showing an even larger increase of 0.087. Grid textures are highly periodic, and defects often manifest as small distortions within a regular lattice. ViT enabled the model to better learn the underlying structure and more accurately isolate local disruptions, a task that conventional convolutional layers struggle with due to their limited receptive field. For the Wood category, the AUC improved from 0.936 to 0.988, a gain of 0.052. This texture exhibits fine-grained directional patterns where defects such as scratches or dents are visually subtle. The ability of ViT to capture long-range dependencies played a crucial role in distinguishing such subtle anomalies. In the Tile category, the improvement was from 0.832 to 0.877, a gain of 0.045. Like Grid, Tile is composed of structured geometric patterns. The ViT module was instrumental in detecting structural irregularities that disrupted these spatial regularities.

Table 3. Performance Comparisons Based on the Presence of ViT Structure Using AUC(↑)

| Dataset | Category | Use ViT X | Use ViT O |
|---|---|---|---|
| MVTec-AD[34] | Carpet | 0.823 | **0.909** |
|  | Grid | 0.886 | **0.973** |
|  | Leather | **0.988** | 0.985 |
|  | Tile | 0.832 | **0.877** |
|  | Wood | 0.936 | **0.988** |
| Surface Crack Detection[35] | Pave&Concrete | 0.847 | **0.856** |
| Marble Surface Anomaly Dataset[36] | Marble | 0.978 | **0.983** |
| Average AUC(↑) |  | 0.899 | **0.939** |

In contrast, the Leather category showed a minor reduction in performance when ViT was used: the AUC slightly decreased from 0.988 (U-Net only) to 0.985 (U-Net-ViT). However, this difference is only 0.003, and both results are within a high-performance range, suggesting that for highly homogeneous and repetitive textures like leather, the addition of global context may offer limited benefit over localized features. The Pave&Concrete category also showed a small but positive improvement, from 0.847 to 0.856, while the Marble category improved from 0.978 to 0.983. These textures are derived from real-world environments and typically contain noise, stains, or irregular damage patterns. Even in these complex categories, the ViT module helped reduce false positives by learning broader contextual structures, enhancing overall reconstruction quality.

In summary, the experimental results clearly demonstrate that the ViT module contributes significantly to the robustness and precision of the defect reconstruction process. The magnitude of improvement varies by texture category, depending on the complexity and regularity of the texture. For unstructured or highly patterned categories such as Carpet, Grid, and Wood, the performance gain is substantial, affirming the necessity of incorporating global information. These results confirm that the hybrid architecture effectively combines frequency-domain enhancement and global spatial understanding, offering a strong advantage over simpler models that rely solely on convolutional features.

*4.3.3 Hyperparameter Analysis*

Hyperparameter Analysis was conducted to analyze the performance of texture-based defect detection using Gabor filters, with variations in filter size and noise probability. The AUC (Area Under the Curve) for each dataset was measured under two key conditions: (1) performance variation with changes in grid size ($k$=16, 8, 4) and (2) performance variation with noise probability ($P_S, P_P$=0, 0.05, 0.1). This study provided a detailed analysis of the optimal filter settings and performance differences across datasets. The grid size plays a significant role in reflecting either the global or local patterns of textures. When noise probability was fixed at $P_S, P_P$=0.05, and grid size ($k$) was varied, the AUC results for each dataset are summarized in Table 4.

Table 4. Performance Comparisons Based on k Settings Using AUC(↑)

| Dataset | Category | k | | |
|---|---|---|---|---|
| | | 16 | 8 | 4 |
| MVTec-AD[34] | Carpet | 0.815 | **0.909** | 0.802 |
| | Grid | 0.970 | **0.973** | 0.928 |
| | Leather | 0.969 | **0.985** | 0.884 |
| | Tile | 0.838 | **0.877** | 0.850 |
| | Wood | 0.967 | **0.988** | 0.951 |
| Surface Crack Detection[35] | Pave&Concrete | 0.863 | 0.856 | **0.865** |
| Marble Surface Anomaly Dataset[36] | Marble | 0.982 | **0.983** | 0.981 |
| Average AUC(↑) | | 0.915 | **0.939** | 0.894 |

In the Carpet category, the highest AUC (0.909) was achieved with $k=8$. This configuration balanced the detection of local defects and the discontinuity between defects and the background. With $k=16$, the boundaries between defects and the background were blurred, resulting in lower performance. Similarly, with $k=4$, the excessive focus on local features diminished the overall defect detection performance. For the Grid category, which features periodic and structured patterns with defects typically appearing as discontinuities in these patterns, $k=8$ achieved the highest AUC (0.973). The grid size $k=16$ resulted in weaker defect region emphasis, while $k=4$ failed to adequately capture the global patterns of the texture, leading to reduced performance. In the Leather category, $k=8$ achieved the highest AUC (0.985), as it effectively captured both the localized defects and subtle texture variations. In contrast, $k=16$ overly emphasized global information, resulting in defect blurring, while $k=4$ overemphasized noisy components of the texture, leading to reduced detection accuracy. The Tile category, which contains symmetrical patterns with small localized defects, achieved its highest AUC (0.877) with $k=8$. This configuration successfully balanced the capture of periodic texture patterns and localized defects. At $k=4$, the algorithm overly focused on localized features, weakening the distinction between defects and the background. At $k=16$, the emphasis on global features hindered defect detection. For the Wood texture, which features wood grain patterns with localized defects, $k=8$ achieved the highest AUC (0.988). This configuration effectively balanced the detection of global patterns and localized defects. With $k=4$, the method emphasized minor variations in wood grain without capturing the broader context, while $k=16$ blurred the defects, leading to a decline in performance. In the Pave&Concrete category, which features irregular and coarse textures with defects such as cracks or surface damage, $k=4$ achieved the highest AUC (0.865). This grid size effectively highlighted the strong localized characteristics of the defects. However, at $k=8$ and $k=16$, the detection performance was relatively lower due to insufficient emphasis on localized patterns. For the Marble texture, which features smooth and glossy characteristics with small defects like cracks and stains, $k=8$ achieved the highest AUC (0.983). While $k=16$ highlighted global features, it did not show significant performance degradation. Meanwhile, $k=4$ successfully captured localized defects but showed greater sensitivity to background noise. Overall, the Marble category exhibited minimal variation in performance across different grid sizes, suggesting a relatively uniform response to grid size changes.

The second ablation study fixed the grid size at $k=8$ while varying noise probabilities at 0, 0.05, and 0.1 to compare AUC (Area Under the Curve) values. The results are summarized in Table 5, demonstrating that a moderate level of noise enhances the boundary distinction between defects and the background. In the Carpet category, the highest performance (AUC 0.909) was achieved with $P_S, P_P=0.05$. This category showed the most significant performance difference compared to other categories, as the optimal noise level effectively emphasized the boundary between textures and defects. At $P_S, P_P=0$, the boundary between textures and defects was indistinct, leading to lower performance, while at $P_S, P_P=0.1$, excessive noise made the training process more challenging. For the Grid category, the highest performance (AUC 0.973) was recorded at $P_S, P_P=0.05$, where noise effectively highlighted the discontinuities in the texture pattern. Although performance remained relatively high at $P_S, P_P=0.1$, excessive noise blurred some defect boundaries. In the Leather category, the highest performance (AUC 0.985) was observed at $P_S, P_P$=0.05. A moderate noise level contributed to accentuating fine texture variations, whereas excessive noise at $P_S, P_P$=0.1 caused defect boundaries to become indistinct, reducing detection accuracy. The Tile and Wood datasets also achieved the highest performance at $P_S, P_P$=0.05, demonstrating that appropriate noise levels simultaneously enhanced texture patterns and defect boundaries. For the Pave&Concrete category, a slight performance increase was observed at $P_S, P_P$=0.1, likely due to the combination of irregular textures and noise accentuating defect boundaries—a unique case in this dataset. In the Marble category, the highest performance (AUC 0.983) was recorded at both $P_S, P_P$=0 and $P_S, P_P$=0.05, with comparable performance at $P_S, P_P$=0.1. This indicates that the texture characteristics of Marble were less sensitive to varying noise levels.

Table 5. Performance Comparisons Based on Noise Probability Settings Using AUC(↑)

| Dataset | Category | $P_S, P_P$ | | |
|---|---|---|---|---|
| | | 0 | 0.05 | 0.1 |
| MVTec-AD[34] | Carpet | 0.841 | **0.909** | 0.803 |
| | Grid | 0.958 | **0.973** | 0.963 |
| | Leather | 0.949 | **0.985** | 0.931 |
| | Tile | 0.867 | 0.877 | **0.848** |
| | Wood | 0.986 | **0.988** | 0.960 |
| Surface Crack Detection[35] | Pave& Concrete | 0.861 | 0.856 | **0.884** |
| Marble Surface Anomaly Dataset[36] | Marble | **0.983** | **0.983** | 0.980 |
| Average AUC(↑) | | 0.921 | **0.939** | 0.910 |

Overall, the experimental results showed that the combination of a filter size of $k$=8 and a noise probability of $P_S, P_P$=0.05 achieved the highest average AUC (0.939). This confirms that appropriate filter size and noise levels positively influence defect detection performance. Moreover, the ability to adaptively configure filter parameters based on the texture characteristics and defect types of each dataset was critical in optimizing performance. The optimization of filter size and noise probability plays a critical role in enhancing defect detection performance. With a filter size of $k$=8, the highest average AUC (0.939) was achieved, while a moderate noise probability of $P_S, P_P$=0.05 effectively accentuated defect patterns. These results indicate the necessity of finely tuning filter size and noise levels to suit the specific conditions of various industrial environments. In particular, it underscores the importance of optimizing filter parameters to adapt to on-site conditions, such as noise levels caused by lighting and equipment variations in manufacturing or inspection processes.

## 5. Discussion

This study proposes a novel framework that seamlessly integrates the traditional signal processing technique of Gabor filtering with a deep learning-based U-Net-ViT hybrid reconstruction network to address the challenges of texture-based defect detection. The proposed method provides an effective solution to common issues in texture images, such as noise-induced responses, ambiguous defect boundaries, and difficulty distinguishing between defective and non-defective regions. By combining the directional and periodic texture analysis capabilities of Gabor filters in the frequency domain with the local-global contextual learning capabilities of the ViT-based reconstruction network in the spatial domain, the framework achieves a structure where defect regions are distinctly highlighted while normal regions are accurately reconstructed.

Gabor filters are known for their sensitivity to specific orientations and frequency components, making them effective for emphasizing structural characteristics of defects. However, in practical applications, they often respond strongly to normal texture patterns as well, resulting in frequent false positives and noise. To overcome this limitation, this study utilizes Gabor filters not as a standalone detection tool but as a preprocessing step. Multi-directional Gabor responses are used to selectively enhance potentially defective regions. These enhanced features are then passed into the U-Net-ViT reconstruction network, which is designed to restore normal patterns while leaving anomalous defect regions unreconstructed. This is not merely a technical combination of methods; rather, it is a goal-driven architecture based on a pipeline of global context reconstruction → frequency-domain enhancement → defect isolation. The framework was experimentally validated to yield robust and accurate defect detection. In particular, this study introduces the SP Masking technique to further improve the clarity of defect representation and guide the learning of the reconstruction network. SP masking introduces artificial noise resembling defects (e.g., black-and-white pixel corruption) into normal images during training. This forces the model to learn how to recover from defect-like perturbations and to clearly distinguish between defective and non-defective areas. By failing to reconstruct only the corrupted regions, the model effectively localizes actual defects. SP noise mimics the randomness and locality of real-world defects, thus enhancing the robustness of the model under various industrial conditions. Furthermore, the framework includes a systematic parameter search and optimization for the core Gabor filter parameters (kernel size, sigma, wavelength, and gamma) based on the unique texture characteristics of each category. Unlike previous approaches that applied generic filtering, this tailored optimization leads to frequency responses that are better suited to the data, providing a meaningful integration of signal processing knowledge with deep learning. This design-driven integration, rather than a simple parallel application, demonstrates both academic and practical value.

Experimental results were validated on seven texture categories, including five from the MVTec AD dataset (Carpet, Leather, Wood, Grid, Tile), and the Pave&Concrete and Marble categories. The proposed method achieved an average AUC of 0.939, significantly outperforming conventional CNN-based reconstruction and standalone Gabor filtering techniques. Notably, for regular textures like Carpet and Grid, the ViT's global contextual modeling played a decisive role in identifying defects, while in irregular textures like Pave and Marble, the strong reconstruction capabilities led to high accuracy. While the integration of U-Net and ViT has been explored in prior research, this study distinguishes itself by incorporating Gabor filtering as a preprocessing stage and leveraging SP masking to simulate defect-like noise during training. The model's strength lies in its deliberate structural integration, where frequency-based emphasis and context-aware reconstruction are jointly optimized. The performance was quantitatively evaluated across diverse texture categories, demonstrating significant improvement. The effectiveness of the proposed architecture is further supported by the results of the ablation study (see Table 3), which compares the U-Net-only structure to the integrated ViT structure. For instance, the AUC score for Carpet improved from 0.823 to 0.909, and for Grid from 0.886 to 0.973 after incorporating ViT. These results confirm that integrating ViT significantly enhances defect detection performance, particularly for regular textures where global context is critical, and for subtle defect patterns requiring fine-grained understanding.

In conclusion, this study presents a powerful architecture that strategically combines Gabor filter-based defect enhancement, U-Net-ViT-based reconstruction, and SP masking-based defect simulation. Together, these components allow for both effective noise suppression and precise defect localization. This framework clearly outperforms conventional or naive combinations and demonstrates high scalability and applicability in texture-based defect detection. The approach has strong potential for real-world deployment in manufacturing, quality inspection, and high-precision visual recognition systems. From a scientific perspective, the integration of classic frequency-domain filtering with neural reconstruction establishes a new and valuable research direction in the field.

## 6. Conclusion

This study presented a novel approach combining Gabor filter-based texture analysis with a Blurring U-Net-ViT hybrid model to achieve robust defect detection in various industrial texture datasets. The Gabor filter effectively emphasized defect regions by responding to specific directional and frequency components, while the reconstruction-based U-Net-ViT model contributed to removing residual noise and restoring defect boundaries with high precision. This integration improved both the reliability and accuracy of detection compared to conventional filtering methods.

Despite its strengths, the proposed approach has several limitations. First, the need to manually search for optimal Gabor filter parameters introduces inefficiency, especially when adapting to new datasets. Second, the U-Net-ViT model, while effective, is computationally intensive, posing challenges for real-time deployment in resource-constrained industrial environments. Third, the method may show reduced performance when dealing with highly complex textures or extremely subtle and overlapping defects.

To address these limitations, future research should consider several directions. One key direction is integrating trainable Gabor modules into an end-to-end architecture, enabling automatic parameter optimization during learning. Additionally, improving robustness under varying acquisition conditions through domain generalization or advanced data augmentation techniques will further enhance applicability. Incorporating multi-scale and attention-based architectures can boost performance in scenarios involving complex or multiple defects. Finally, model compression and computational optimization are essential steps to ensure real-time performance for deployment in industrial inspection systems.